\documentclass[11pt,letterpaper]{article}
\usepackage{emnlp2016}
\usepackage{subcaption}
\usepackage[font=footnotesize,labelfont=bf]{caption}
\usepackage{amssymb}
\usepackage{times}
\usepackage{booktabs}
\usepackage{latexsym}
\usepackage{graphicx}
\usepackage{multirow}
\usepackage{enumitem}
\usepackage{amsmath}
\usepackage{url}

\emnlpfinalcopy

\def\emnlppaperid{***}

\newcommand{\tbf}{\textbf}
\newcommand{\sentprop}{{\sc{SentProp}}} 
\newcommand{\densify}{{\sc{Densifier}}} 
\newcommand{\socsent}{{\sc{SocialSent}} }
\newcommand{\mb}{\mathbf} 
\newcommand{\R}{\mathbb{R}}

\makeatletter
\def\citealt{\def\citename##1{{\frenchspacing##1}, }\@internalcitec}

\def\@citexc[#1]#2{\if@filesw\immediate\write\@auxout{\string\citation{#2}}\fi
  \def\@citea{}\@citealt{\@for\@citeb:=#2\do
    {\@citea\def\@citea{;\penalty\@m\ }\@ifundefined
       {b@\@citeb}{{\bf ?}\@warning
       {Citation `\@citeb' on page \thepage \space undefined}}%
{\csname b@\@citeb\endcsname}}}{#1}}

\def\@internalcitec{\@ifnextchar [{\@tempswatrue\@citexc}{\@tempswafalse\@citexc[]}}

\def\@citealt#1#2{{#1\if@tempswa, #2\fi}}
\makeatother

\title{Inducing Domain-Specific Sentiment Lexicons from Unlabeled Corpora}

\author{William L.\@ Hamilton, Kevin Clark, Jure Leskovec, Dan Jurafsky \\
Department of Computer Science, Stanford University, Stanford CA, 94305\\
\texttt{wleif,kevclark,jure,jurafsky@stanford.edu}}

\date{}

\begin{document}
\maketitle

\begin{abstract}
A word's sentiment depends on the domain in which it is used. 
Computational social science research thus requires sentiment lexicons that are specific to the domains being studied.
We combine domain-specific word embeddings with a label propagation framework to induce accurate domain-specific sentiment lexicons using small sets of seed words, achieving state-of-the-art performance competitive with approaches that rely on hand-curated resources.
Using our framework we perform two large-scale empirical studies to quantify the extent to which sentiment varies across time and between communities. 
We induce and release historical sentiment lexicons for 150 years of English and community-specific sentiment lexicons for 250 online communities from the social media forum Reddit. 
The historical lexicons show that more than $5\%$ of sentiment-bearing (non-neutral) English words completely switched polarity during the last 150 years, and the community-specific lexicons highlight how sentiment varies drastically between different communities. 
\end{abstract}

\section{Introduction}
\begin{figure}
\centering
\includegraphics[scale=1.0]{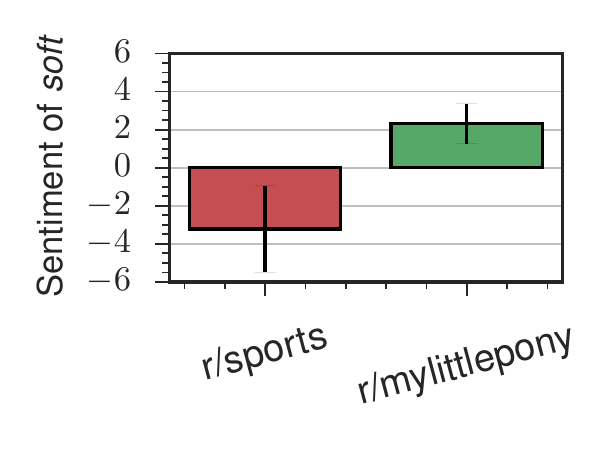}
\vspace{-10pt}
\caption{{\footnotesize \textbf{The sentiment of \textit{soft} in different online communities.} Sentiment values computed using \sentprop\ (Section 3) on comments from Reddit communities illustrate how sentiment depends on social context. Bootstrap-sampled standard deviations provide a measure of confidence with the scores.}}
\label{soft}
\vspace{-15pt}
\end{figure}

\begin{figure}
\centering
\includegraphics[width=0.9\columnwidth]{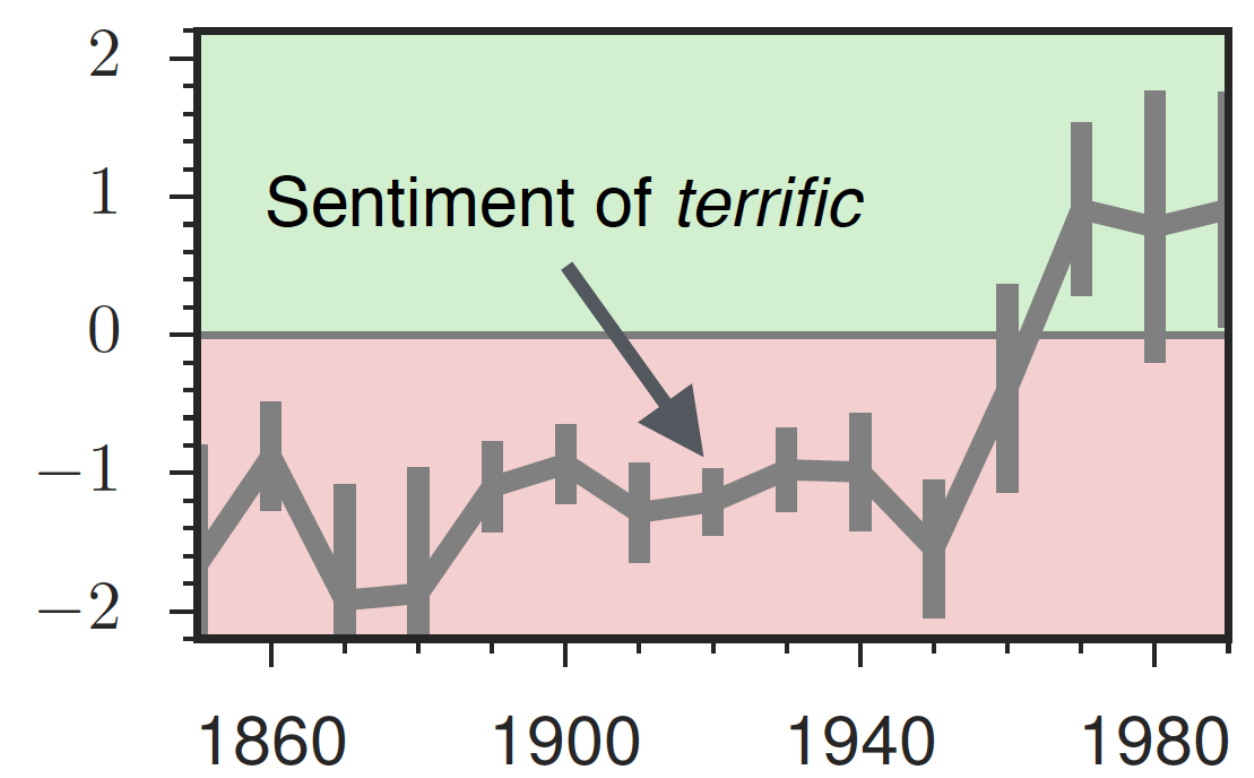}
\vspace*{-0pt}
\caption{{\footnotesize\textbf{The sentiment of \textit{terrific} changed from negative to positive over the last 150 years}. Sentiment values and bootstrapped confidences were computed using \sentprop\ on historical data (see Section \ref{historical}).}}
\label{terrific}
\vspace{-25pt}
\end{figure}

Inducing domain-specific sentiment lexicons is crucial to computational social science (CSS) research.
Sentiment lexicons allow us to analyze key subjective properties of texts like opinions and attitudes \cite{taboada_lexicon-based_2011}.
But lexical  sentiment is hugely influenced by context.
The word \textit{soft} has a very different sentiment in an online sports community 
than it does in one dedicated to toy animals (Figure \ref{soft}).
\textit{Terrific} once had a highly negative connotation; now it is essentially synonymous with \textit{good} (Figure \ref{terrific}).
Without domain-specific lexicons, social scientific analyses can be misled by sentiment assignments 
biased towards domain-general contexts, neglecting factors like
genre, community-specific vernacular, or demographic variation \cite{deng2014joint,hovy_demographic_2015,yang_putting_2015}.

Using experts or crowdsourcing to construct domain-specific sentiment lexicons is expensive and often
time-consuming \cite{mohammad_emotions_2010,fast_empath:_2016}, and is 
especially problematic when non-standard language (as in historical documents or obscure social media forums)
prevents annotators from understanding the sociolinguistic context of the data.
 
Web-scale sentiment lexicons can be automatically induced for large socially-diffuse domains, such as the internet-at-large \cite{velikovich_viability_2010} or all of Twitter \cite{tang_building_2014}.
However, to study sentiment in domain-specific cases---financial documents, historical texts, or tight-knit social media forums---such  generic lexicons  may be inaccurate,  and  even introduce harmful biases \cite{loughran_when_2011}.\footnote{\tiny \url{http://brandsavant.com/brandsavant/the-hidden-bias-of-social-media-sentiment-analysis}}
Researchers need a principled and accurate framework for inducing lexicons that are specific to their domain of study. 

To meet these needs, we introduce \sentprop, a framework to learn accurate sentiment lexicons from small sets of seed words and domain-specific corpora. 
\sentprop\ combines the well-known method of label propagation with advances in word embeddings, and 
unlike previous approaches, is designed to be accurate even when using modestly-sized domain-specific corpora (${\sim}10^7$ tokens). 
Our framework also provides {\em confidence scores} along with the learned lexicons, which allows researchers to quantify uncertainty in a principled manner.

The key contributions of this work are: 
\begin{enumerate}[leftmargin=*, topsep=2pt, itemsep=0pt, parsep=2pt]
\item
A simple state-of-the-art sentiment induction algorithm, combining high-quality word vector embeddings with a label propagation approach.
\item
A novel bootstrap-sampling framework for inferring confidence scores with the sentiment values. 
\item
Two large-scale studies that reveal how sentiment depends on both social and historical context.\\
(a)~ We induce community-specific sentiment lexicons for the largest 250 ``subreddit'' communities on the social-media forum Reddit, revealing substantial variation in word sentiment between communities. \\
(b)~ We induce historical sentiment lexicons for 150 years of English, revealing that ${>}5\%$ of words switched polarity during this time.
\end{enumerate}
To the best of our knowledge, this is the first work to systematically analyze the domain-dependency of sentiment at a large-scale, across hundreds of years and hundreds of user-defined online communities.

All of the inferred lexicons along with code for \sentprop\ and all methods evaluated are made available in the \socsent package released with this paper.\footnote{\scriptsize\url{http://nlp.stanford.edu/projects/socialsent}}
The \socsent\ package provides a benchmark toolkit for inducing sentiment lexicons, including implementations of previously published algorithms \cite{velikovich_viability_2010,rothe_ultradense_2016}, which are not otherwise publicly available. 

\section{Related work}
\begin{table*}[t!]
\centering
{\fontsize{10}{11}\selectfont
\begin{tabular}{p{0.5in}p{2.5in}p{2.5in}}
\toprule  
Domain & Positive seed words & Negative seed words\\ 
\midrule 
Standard English & good, lovely, excellent, fortunate, pleasant, delightful, perfect, loved, love, happy & bad, horrible, poor, unfortunate, unpleasant, disgusting, evil, hated, hate, unhappy \\
\addlinespace[3pt]
Finance & successful, excellent, profit, beneficial, improving, improved, success, gains, positive & negligent, loss, volatile, wrong, losses, damages, bad, litigation, failure, down, negative  \\
\addlinespace[3pt]
Twitter & love, loved, loves, awesome, nice, amazing, best, fantastic, correct, happy & hate, hated, hates, terrible, nasty, awful, worst, horrible, wrong, sad\\
\bottomrule
\end{tabular} 
}
\caption{{\footnotesize\textbf{Seed words.} The seed words were manually selected to be context insensitive (without knowledge of the test lexicons).}}
\label{seeds}
\vspace{-10pt}
\end{table*}

Our work builds upon a wealth of previous research on inducing sentiment lexicons, along two threads:

{\em Corpus-based} approaches use seed words and patterns in unlabeled corpora to induce domain-specific lexicons.
These patterns may rely on syntactic structures 
\cite{hatzivassiloglou_predicting_1997,jijkoun_generating_2010,rooth_inducing_1999,thelen_bootstrapping_2002,widdows_graph_2002}, which can be domain-specific and brittle (e.g., in social media lacking usual grammatical structures).
 Other models rely on general co-occurrence \cite{igo_corpus-based_2009,riloff_corpus-based_1997,turney_measuring_2003}. 
Often corpus-based methods exploit distant-supervision signals (e.g., review scores, emoticons) specific to certain domains \cite{asghar_unified_2015,blair-goldensohn_building_2008,bravo-marquez_unlabelled_2015,choi_adapting_2009,severyn_automatic_2015,speriosu_twitter_2011,tang_building_2014}.
An effective corpus-based approach that does not require distant-supervision---which we adapt here---is to construct lexical graphs using word co-occurrences and then to perform some form of label propagation over these graphs \cite{huang_automatic_2014,velikovich_viability_2010}.
Recent work has also learned transformations of word-vector representations in order to induce sentiment lexicons \cite{rothe_ultradense_2016}.
\newcite{fast_empath:_2016} combine word vectors with crowdsourcing to produce domain-independent topic lexicons.

{\em Dictionary-based} approaches use hand-curated lexical resources---usually WordNet \cite{fellbaum_wordnet_1998}---in order to propagate sentiment from seed labels \cite{esuli_sentiwordnet:_2006,hu_mining_2004,kamps_using_2004,rao_semi-supervised_2009,san_vicente_simple_2014,takamura_extracting_2005,tai_automatic_2013}.
There is an implicit consensus that dictionary-based approaches will generate higher-quality lexicons, due to their use of these clean, hand-curated resources; however, they are not applicable in domains lacking such a resource (e.g., most historical texts). 

Most previous work seeks to enrich or enlarge existing lexicons \cite{qiu_expanding_2009,san_vicente_simple_2014,velikovich_viability_2010}, emphasizing recall over precision. 
This recall-oriented approach is motivated by the need for massive polarity lexicons in tasks like web-advertising \cite{velikovich_viability_2010}.
 In contrast to these previous efforts, the goal of this work is to induce high-quality lexicons that are accurate to a specific social context.

Algorithmically, our approach is inspired by \newcite{velikovich_viability_2010}. We extend \newcite{velikovich_viability_2010} by incorporating high-quality word vector embeddings, a new graph construction approach, an alternative label propagation algorithm, and a bootstrapping method to obtain confidence values. 
Together these improvements, especially the high-quality word vectors, allow our corpus-based method to even outperform the state-of-the-art dictionary-based approach.

\section{Framework}\label{framework}
\begin{figure*}
\centering
\includegraphics[width=1.\textwidth]{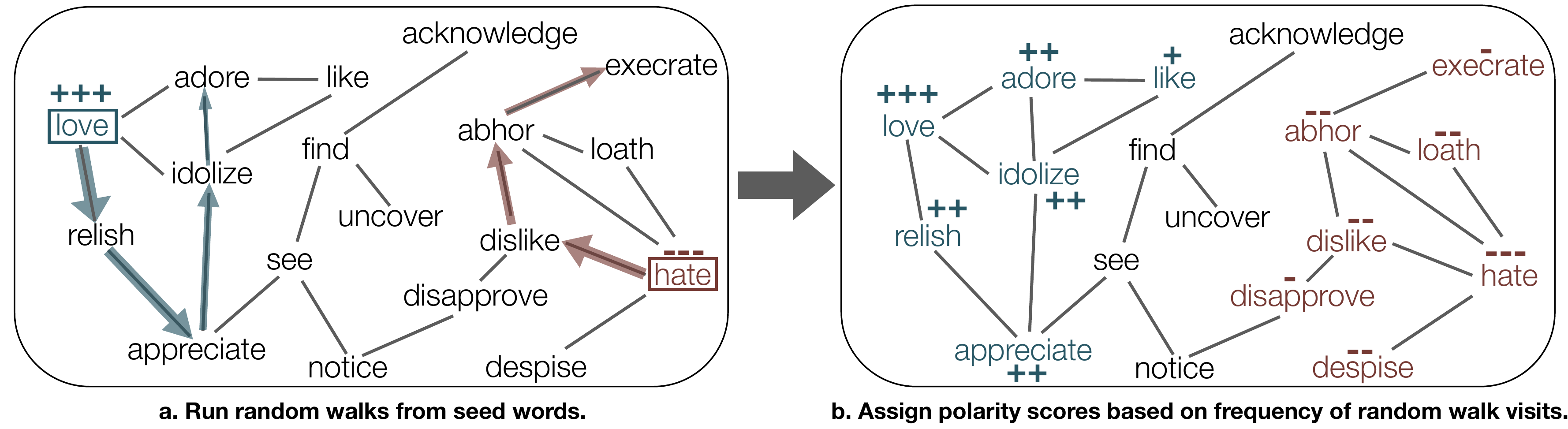}
\caption{Visual summary of the \sentprop\ algorithm.}
\vspace{-10pt}
\label{diagram}
\end{figure*}

Our framework, \sentprop, is designed to meet four key desiderata:
\begin{enumerate}[itemsep=0pt,topsep=1pt, parsep=0pt, leftmargin=*]
\item
	\textbf{Resource-light:} Accurate performance without massive corpora or hand-curated resources. 
\item
	\textbf{Interpretable}: Uses small seed sets of ``paradigm'' words to maintain interpretability and avoid ambiguity in sentiment values.
\item
	\textbf{Robust}: Bootstrap-sampled standard deviations provide a measure of confidence.
\item
	\textbf{Out-of-the-box}: Does not rely on signals that are specific to only certain domains. 
\end{enumerate}

\sentprop\ involves two steps: constructing a lexical graph from unlabeled corpora and propagating sentiment labels over this graph.

\subsection{Constructing a lexical graph}

Lexical graphs are constructed from distributional word embeddings learned on unlabeled corpora.  

\subsubsection*{Distributional word embeddings}

The first step in our approach is to build high-quality semantic representations for words using a vector space model (VSM).  
We embed each word $w_i \in \mathcal{V}$ as a vector $\mathbf{w}_i$ that captures information about its co-occurrence statistics with other words \cite{landauer_solution_1997,turney_frequency_2010}.
This  VSM approach has a long history in NLP and has been highly successful in recent applications (see \citealt{levy_improving_2015} for a survey). 

When recreating known lexicons, we used a number of publicly available embeddings (Section \ref{evaluation}). 

In the cases where we learned embeddings ourselves, we employed an SVD-based method to construct the word-vectors.
First, we construct a matrix $\mb{M}^{\sc{PPMI}} \in \R^{|\mathcal{V}| \times |\mathcal{V}|}$ with entries given by
\begin{equation}
\mb{M}^{\sc{PPMI}}_{i,j} = \max\left\lbrace\log\left(\frac{\hat{p}(w_i,w_j)}{\hat{p}(w)\hat{p}(w_j)}\right) ,0\right\rbrace,
\end{equation}
 where $\hat{p}$ denotes smoothed empirical probabilities of word \mbox{(co-)occurrences} within fixed-size sliding windows of text.\footnote{We use contexts of size four on each side and context-distribution smoothing with $c=0.75$ \cite{levy_improving_2015}.}
$\mb{M}^{\sc{PPMI}}_{i,j}$ is equal to a smoothed variant of the positive pointwise mutual information between words $w_i$ and $w_j$ \cite{levy_improving_2015}. 
Next, we compute $\mb{M}^{\sc{PPMI}} = \mb{U}\mb{\Sigma}\mb{V}^\top$, the truncated singular value decomposition of $\mb{M}^{\sc{PPMI}}$. 
The vector embedding for word $w_i$ is then given by 
\begin{equation}
\mb{w}^{\textrm{SVD}}_{i} = \left(\mb{U}\right)_{i}.
\end{equation}
Excluding the singular value weights, $\Sigma$,  has been shown known to dramatically improve embedding quality \cite{turney_frequency_2010,bullinaria_extracting_2012}. 
 Following standard practices, we learn embeddings of dimension 300. 
 
 We found that this SVD-based method significantly outperformed word2vec \cite{mikolov_distributed_2013} and GloVe \cite{pennington_glove:_2014} on preliminary experiments with the domain-specific data we used (see Section \ref{evaluation}).

\subsubsection*{Defining the graph edges}

Given a set of word embeddings, a weighted lexical graph is constructed by connecting each word with its nearest $k$ neighbors within the semantic space (according to cosine-similarity).
The weights of the edges are set as
\begin{equation}\label{edgeweights}
\mb{E}_{i,j} = \arccos\left(-\frac{\mathbf{w_i}^\top\mathbf{w_j}}{\|\mathbf{w_i}\|\|\mathbf{w_j}\|}\right).
\end{equation}

\subsection{Propagating polarities from a seed set}

Once a weighted lexical graph is constructed, we propagate sentiment labels over this graph using a random walk method \cite{zhou_learning_2004}.
A word's polarity score for a seed set is proportional to the probability of a random walk from the seed set hitting that word (Figure \ref{diagram}). 

Let  $\mb{p} \in \R^{|\mathcal{V}|}$ be a vector of word-sentiment scores constructed using seed set $\mathcal{S}$ (e.g., ten negative words); $\mb{p}$ is initialized to have $\frac{1}{|\mathcal{V}|}$ in all entries.  
And let $\mb{E}$ be the matrix of edge weights given by equation \eqref{edgeweights}. 
First, we construct a symmetric transition matrix from $\mb{E}$ by computing $
\mb{T} = \mb{D}^{\frac{1}{2}}\mb{E}\mb{D}^{\frac{1}{2}}$,
where $\mb{D}$ is a matrix with the column sums of $\mb{E}$ on the diagonal. 
Next, using $\mb{T}$ we iteratively update $\mb{p}$ until numerical convergence:
\begin{equation}
\mb{p}^{(t+1)} = \beta\mb{T}\mb{p}^{(t)} + (1-\beta)\mb{s},
\end{equation}
where $\mb{s}$ is a vector with values set to $\frac{1}{|\mathcal{S}|}$ in the entries corresponding to the seed set $\mathcal{S}$ and zeros elsewhere. 
The $\beta$ term controls the extent to which the algorithm favors local consistency (similar labels for neighbors) vs. global consistency (correct labels on seed words), with lower $\beta$s emphasizing the latter. 

To obtain a final polarity score for a word $w_i$, we run the walk using both positive and negative seed sets, obtaining positive ($\mb{p}^{P}(w_i)$) and negative ($\mb{p}^{N}(w_i)$) label scores. We then combine these values into a positive-polarity score as $\bar{\mb{p}}^P(w_i) = \frac{\mb{p}^{P}(w_i)}{\mb{p}^{P}(w_i) + \mb{p}^{N}(w_i)}$ and standardize the final scores to have zero mean and unit variance (within a corpus). 

\subsection{\sentprop\ variants}

Many variants of the random walk approach and related label propagation techniques exist in the literature \cite{san_vicente_simple_2014,velikovich_viability_2010,zhou_learning_2004,zhu_learning_2002,zhu_semi-supervised_2003}.
For example, there are differences in how to normalize the transition matrix in the random walks \cite{zhou_learning_2004} and variants of label propagation, e.g. where the labeled seeds are clamped to the correct values \cite{zhu_learning_2002} or where only shortest-paths through the graph are used for propagation \cite{velikovich_viability_2010}.

We experimented with a number of these approaches and found little difference in their performance.
We opted to use the random walk method because it had a slight edge in terms of performance in preliminary experiments\footnote{${>}2\%$ improvement across metrics on the standard and historical English datasets described in Section \ref{evaluation}.} and because it produces well-behaved distributions over label scores, whereas \newcite{zhu_learning_2002}'s method and its variants produce extremely peaked distributions. 
We do note report in detail on all the label propagation variants here, but the \socsent\ package contains a full suite of these methods.

\subsection{Bootstrap-sampling for robustness}

Propagated sentiment scores are inevitably influenced by the seed set, and it is important for researchers to know the extent to which polarity values are simply the result of corpus artifacts that are correlated with these seeds words. 
We address this issue by using a bootstrap-sampling approach to obtain confidence regions over our sentiment scores. 
We bootstrap by running our propagation over $B$ random equally-sized subsets of the positive and negative seed sets.
Computing the standard deviation of the bootstrap-sampled polarity scores provides a measure of confidence and allows the researcher to evaluate the robustness of the assigned polarities.
We set $B=50$ and used 7 words per random subset (full seed sets are size 10; see Table \ref{seeds}).  

\section{Recreating known lexicons}\label{evaluation}

We validate our approach by recreating known sentiment lexicons in the three domains: Standard English, Twitter, and Finance. Table \ref{seeds} lists the seed words used in each domain. 

{\bf Standard English:}~ To facilitate comparison with previous work, we focus on the well-known General Inquirer lexicon \cite{stone_general_1966}. We also use the continuous valence (i.e., polarity) scores collected by \newcite{warriner_norms_2013} in order to evaluate the fine-grained performance of our framework. 
We test our framework's performance using two different embeddings: off-the-shelf Google news embeddings constructed from $10^{11}$ tokens\footnote{\url{https://code.google.com/p/word2vec/}} and embeddings we constructed from the 2000s decade of the Corpus of Historical American English (COHA), which contains ${\sim}2\times10^7$ words in each decade, from 1850 to 2000 \cite{davies_corpus_2010}. 
The COHA corpus allows us to test how the algorithms deal with this smaller historical corpus, which is important since we will use the COHA corpus to infer historical sentiment lexicons (Section \ref{historical}).

{\bf Finance:} Previous work found that general purpose sentiment lexicons performed very poorly on financial text \cite{loughran_when_2011}, so a finance-specific sentiment lexicon (containing binary labels) was hand-constructed for this domain (ibid.). 
To test against this lexicon, we constructed embeddings using a dataset of ${\sim}2\times10^7$ tokens from financial 8K documents \cite{lee_importance_2014}.

{\bf Twitter:} Numerous works attempt to induce Twitter-specific sentiment lexicons using supervised approaches and features unique to that domain (e.g., follower graphs; \citealt{speriosu_twitter_2011}). Here, we emphasize that we can induce an accurate lexicon using a simple domain-independent and resource-light approach, with the implication that lexicons can easily be induced for related social media domains without resorting to complex supervised frameworks. 
We evaluate our approach using the test set from the 2015 SemEval task 10E competition \cite{rosenthal_semeval-2015_2015}, and we use the embeddings constructed by \newcite{rothe_ultradense_2016}.\footnote{The official SemEval task 10E involved fully-supervised learning, so we do not use their evaluation setup.}

\subsection{Baselines and state-of-the-art comparisons}
\begin{table*}
\begin{subtable}{0.47\textwidth}
\centering
{\small
\begin{tabular}{lccc} 
\toprule  
Method & AUC & Ternary F1 & $\tau$\\ 
\midrule 
\sentprop & 90.6 & 58.6 & 0.44\\
\densify & \tbf{93.3} & \tbf{62.1} & \tbf{0.50}\\
WordNet & 89.5 & 58.7 & 0.34\\
Majority  & -- & 24.8 & --\\
\bottomrule
\end{tabular} 
}
\caption{{\footnotesize\textbf{Corpus methods outperform WordNet on standard English}. Using word-vector embeddings learned on a massive corpus ($10^{11}$ tokens), we see that both corpus-based methods outperform the WordNet-based approach overall.}}
\label{english}
\end{subtable}
\hfill
\begin{subtable}{0.47\textwidth}
\centering
{\small
\begin{tabular}{lccc} 
\toprule  
Method & AUC & Ternary F1 & $\tau$\\ 
\midrule 
\sentprop & 86.0 & \tbf{60.1} & 0.50\\
\densify & \tbf{90.1} & 59.4 & \tbf{0.57}\\
Sentiment140 & 86.2 & 57.7 & 0.51\\
Majority & -- & 24.9 & -- \\
\bottomrule
\end{tabular} 
}
\caption{{\footnotesize\textbf{Corpus approaches are competitive with a distantly supervised method on Twitter}. Using Twitter embeddings learned from ${\sim}10^9$ tokens, we see that the semi-supervised corpus approaches using small seed sets perform very well.}}
\label{twitter}
\end{subtable}

\vspace{10pt}
\begin{subtable}{0.47\textwidth}
\centering
{\small
\begin{tabular}{lcc} 
\toprule  
Method & AUC & Ternary F1\\ 
\midrule 
\sentprop & \tbf{91.6} & \tbf{63.1}\\
\densify & 80.2 & 50.3\\
PMI & 86.1 & 49.8\\
\newcite{velikovich_viability_2010} & 81.6 & 51.1\\
Majority & -- & 23.6\\
\bottomrule
\end{tabular} 
}
\caption{{\footnotesize\textbf{\sentprop\ performs best with domain-specific finance embeddings.}
Using embeddings learned from financial corpus (${\sim}2\times10^7$ tokens) , \sentprop\ significantly outperforms the other methods.}}
\label{finance}
\end{subtable}
\hfill
\begin{subtable}{0.47\textwidth}
\centering
{\small
\begin{tabular}{lccc} 
\toprule  
Method & AUC & Ternary F1 & $\tau$\\ 
\midrule 
 \sentprop & \tbf{83.8} & \tbf{53.0} & \tbf{0.28}\\
 \densify & 77.4 & 46.6 & 0.19\\
 PMI  & 70.6 & 41.9 & 0.16\\
 \newcite{velikovich_viability_2010}  & 52.7 & 32.9 & 0.01\\
 Majority & -- & 24.3 & --\\
\bottomrule
\end{tabular} 
}
\caption{{\footnotesize \textbf{\sentprop\ performs well on standard English even with 1000x reduction in corpus size.} 
\sentprop\ maintains strong performance even when using embeddings learned from the 2000s decade of COHA (only $2\times{\sim}10^7$ tokens). 
}}
\label{english-small}
\end{subtable}
\caption{\footnotesize\textbf{Results on recreating known lexicons.}}
\vspace{-10pt}
\end{table*}
We compare \sentprop\ against standard baselines and state-of-the-art approaches. 
The PMI baseline of \newcite{turney_measuring_2003} computes the pointwise mutual information between the seeds and the targets without using propagation.
 The baseline method of \newcite{velikovich_viability_2010} is similar to our method but uses an alternative propagation approach and raw co-occurrence vectors instead of learned embeddings. 
 Both these methods require raw corpora, so they function as baselines in cases where we do not use off-the-shelf embeddings.
We also compare against \densify, a state-of-the-art method that learns orthogonal transformations of word vectors instead of propagating labels \cite{rothe_ultradense_2016}.
Lastly, on standard English we compare against a state-of-the-art WordNet-based method, which performs label propagation over a WordNet-derived graph \cite{san_vicente_simple_2014}. Several variant baselines, all of which \sentprop\ outperforms, are omitted for brevity (e.g., using word-vector cosines in place of PMI in \newcite{turney_measuring_2003}'s framework). 
Code for all these variants is available in the \socsent\ package.

\subsection{Evaluation setup}

We evaluate the approaches according to (i) their binary classification accuracy (ignoring the neutral class, as is common in previous work), (ii) ternary classification performance (positive vs.\@ neutral vs.\@ negative)\footnote{Only GI contains words explicitly marked neutral, so for ternary evaluations in Twitter and Finance we sample neutral words from GI to match its neutral-vs-not distribution. }, and (iii) Kendall $\tau$ rank-correlation with continuous human-annotated polarity scores. 

For all methods in the ternary-classification condition, we use the class-mass normalization method \cite{zhu_semi-supervised_2003} to label words as positive, neutral, or negative.
This method assumes knowledge of the label distribution---i.e., how many positive/negative vs. neutral words there are---and simply assigns labels to best match this distribution.

\subsection{Evaluation results}

Tables \ref{english}-\ref{english-small} summarize the performance of our framework along with baselines and other state-of-the-art approaches. 
Our framework significantly outperforms the baselines on all tasks, outperforms a state-of-the-art approach that uses WordNet on standard English (Table \ref{english}), and is competitive with Sentiment140 on Twitter (Table \ref{twitter}), a distantly-supervised approach that uses signals from emoticons \cite{mohammad_emotions_2010}. 
\densify\ also performs extremely well, outperforming \sentprop\ when off-the-shelf embeddings are used (Tables \ref{english} and \ref{twitter}). 
However, \sentprop\ significantly outperforms all other approaches when using the domain-specific embeddings (Tables \ref{finance} and \ref{english-small}). 

Overall our results show that \sentprop --- a relatively simple method, which combines high-quality word vectors embeddings with standard label propagation --- can perform at a state-of-the-art level, even performing competitively with methods relying on hand-curated lexical graphs.
Unlike previous published approaches, \sentprop\ is able to maintain high accuracy even when modest-sized domain-specific corpora are used. 
In cases where very large corpora are available and where there is an abundance of training data, \densify\ performs extremely well, since it was designed for this sort of setting \cite{rothe_ultradense_2016}. 

We found that the baseline method of \newcite{velikovich_viability_2010}, which our method is closely related to, performed relatively poorly with these domain-specific corpora.
This indicates that using high-quality word-vector embeddings can have a drastic impact on performance.
However, it is worth noting that \newcite{velikovich_viability_2010}'s method was designed for high recall with massive corpora, so its poor performance in our regime is not surprising.

Lastly, we found that the choice of embedding method could have a drastic impact. 
Preliminary experiments on the COHA data showed that using word2vec SGNS vectors (with default settings) instead of our SVD-based embeddings led to a ${>}40\%$ performance drop for \sentprop\ across all measures and a ${>}10\%$ performance drop for \densify. 
It is possible that certain settings of word2vec could perform better, but previous work has shown that SVD-based methods have superior results on smaller datasets and rare-word similarity tasks \cite{levy_improving_2015,hamilton_diachronic_2016}, so this result is not surprising.

\section{Inducing community-specific lexicons}

As a first large-scale study, we investigate how sentiment depends on the social context in which a word is used. 
It is well known that there is substantial sociolinguistic variation between different communities, whether these communities are defined geographically \cite{trudgill_linguistic_1974} or via underlying sociocultural differences \cite{labov_social_2006}. 
However, no previous work has systematically investigated community-specific variation in word sentiment at a large scale. 
\newcite{yang_putting_2015} exploit social network structure in Twitter to infer a small number (1-10) of communities and analyzed sentiment variation via a supervised framework. 
Our analysis extends this line of work by analyzing the sentiment across hundreds of user-defined communities using only unlabeled corpora and a small set of ``paradigm'' seed words (the Twitter seed words outlined in Table \ref{seeds}).  

In our study, we induced sentiment lexicons for the top-250 (by comment-count) subreddits from the social media forum Reddit.\footnote{Subreddits are user-created topic-specific forums.} 
We used all the 2014 comment data to induce the lexicons, with words lower cased and comments from bots and deleted users removed.\footnote{\tiny\url{https://archive.org/details/2015_reddit_comments_corpus}}
Sentiment was induced for the top-5000 non-stop words in each subreddit (again, by comment-frequency).

\subsection{Examining the lexicons}

\begin{figure*}[ht!]
\centering
\includegraphics[width=0.95\textwidth]{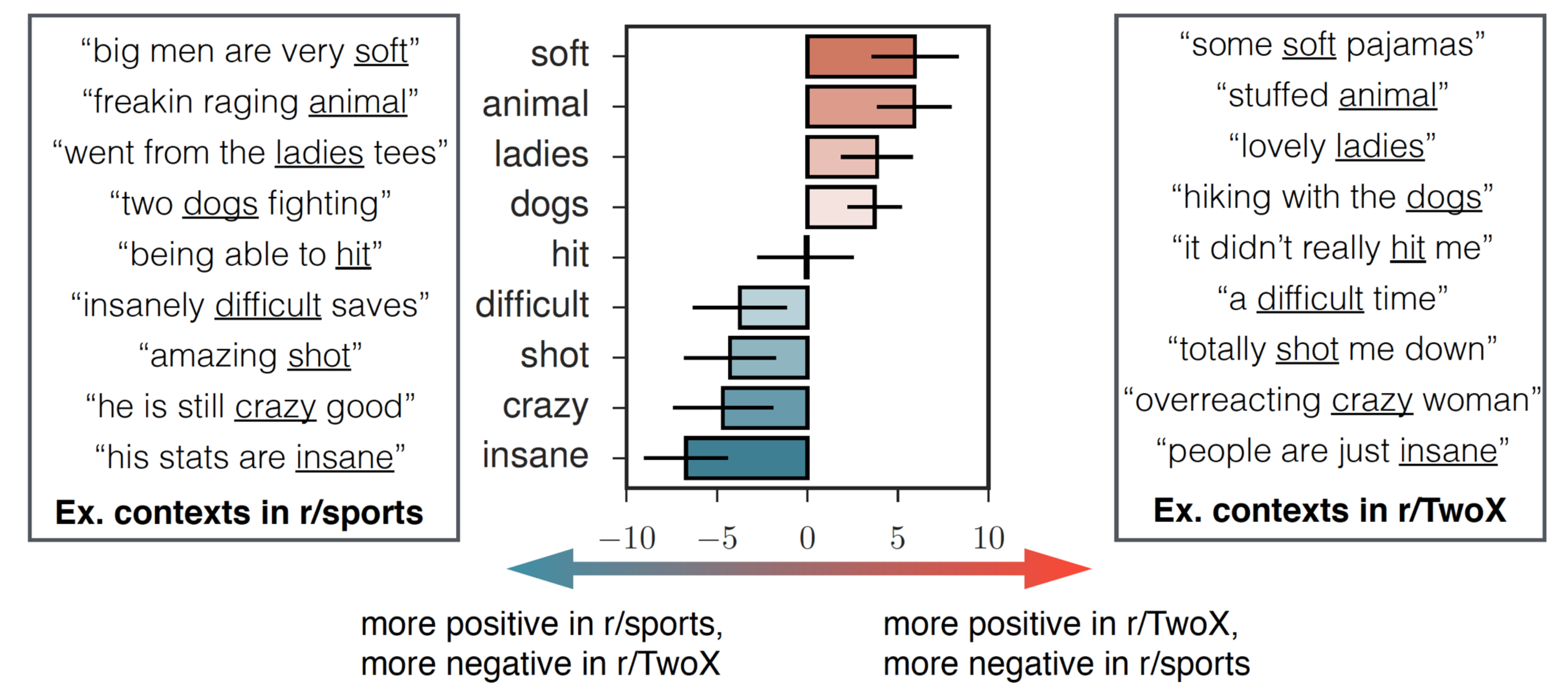}
\vspace{0pt}
\caption{{\footnotesize\textbf{Word sentiment differs drastically between a community dedicated to sports (\texttt{r/sports}) and one dedicated to female perspectives and gender issues (\texttt{r/TwoX}).} Words like \textit{soft} and \textit{animal}  have positive sentiment in \texttt{r/TwoX} but negative sentiment in \texttt{r/sports}, while the opposite holds for words like \textit{crazy} and \textit{insane}}.}
\label{twoxvssports}
\vspace{-15pt}
\end{figure*}

Analysis of the learned lexicons reveals the extent to which sentiment can differ across communities. 
Figure \ref{twoxvssports} highlights some words with opposing sentiment in two communities: in \texttt{r/TwoXChromosomes} (\texttt{r/TwoX}), a community dedicated to female perspectives and gender issues, 
the words \textit{crazy} and \textit{insane} have negative polarity, which is not true in the \texttt{r/sports} community, and, vice-versa, words like \textit{soft} are positive in \texttt{r/TwoX} but negative in \texttt{r/sports}. 

\begin{figure}[t!]
\centering
\includegraphics[width=1.0\columnwidth]{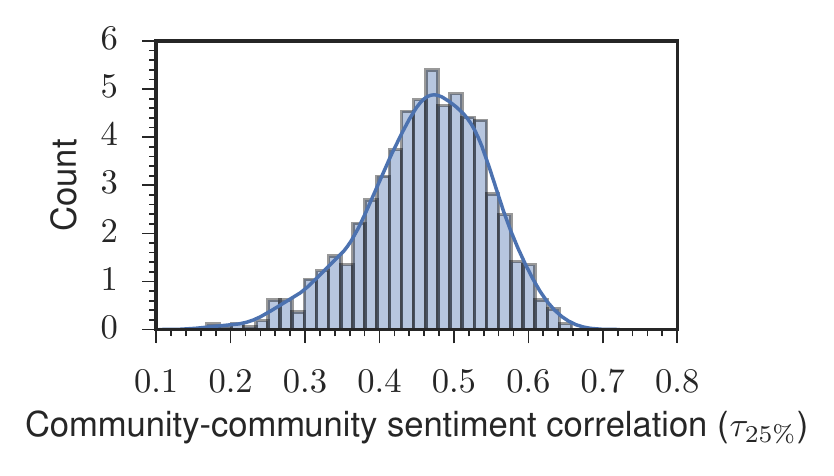}
\vspace{-10pt}
\caption{{\footnotesize\textbf{There is a long tail of communities with very different word sentiments.} Some communities have very similar sentiment (e.g., \texttt{r/sports} and \texttt{r/hockey}), while other community pairs differ drastically (e.g., \texttt{r/sports} and \texttt{r/TwoX}).}}
\vspace{0pt}
\label{community_corr}
\end{figure}

To get a sense of how much sentiment differs across communities in general, we selected a random subset of 1000 community pairs and examined the correlation in their sentiment values for highly sentiment-bearing words (Figure \ref{community_corr}).
We see that the distribution is noticeably skewed, with many community pairs having highly uncorrelated sentiment values. 
The 1000 random pairs were selected such that each member of the pair overlapped in at least half of their top-5000 word vocabulary. 
We then computed the correlation between the sentiments in these community-pairs.
Since sentiment is noisy and relatively uninteresting for neutral words, we compute $\tau_{25\%}$, the Kendall-$\tau$ correlation over the top-25\% most sentiment bearing words shared between the two communities. 


Analysis of individual pairs reveals some interesting insights about sentiment and inter-community dynamics. 
For example, we found that the sentiment correlation between \texttt{r/TwoX} and \texttt{r/TheRedPill} ($\tau_{25\%}=0.58$), two communities that hold conflicting views and often attack each other\footnote{This conflict is well-known on Reddit; for example, both communities mention each others' names along with \textit{fuck}-based profanity in the same comment far more than one would expect by chance ($\chi^2_1 > 6.8$, $p<0.01$ for both). \texttt{r/TheRedPill} is dedicated to male empowerment.},
was actually higher than the sentiment correlation between \texttt{r/TwoX} and \texttt{r/sports} ($\tau_{25\%}=0.41$), two communities that are entirely unrelated. 
This result suggests that conflicting communities may have more similar sentiment in their language compared to communities that are entirely unrelated.

\section{Inducing diachronic sentiment lexicons}\label{historical}

Sentiment also depends on the historical time-period in which a word is used. 
To investigate this dependency, we use our framework to analyze how word polarities have shifted over the last 150 years. 
The phenomena of {\em amelioration} (words becoming more positive) and {\em pejoration} (words becoming more negative) are well-discussed in the linguistic literature \cite{traugott_regularity_2001}; however, no comprehensive polarity lexicons exist for historical data \cite{cook_automatically_2010}. 
Such lexicons are crucial to the growing body of work on NLP analyses of historical text \cite{piotrowski_natural_2012} which are informing diachronic linguistics \cite{hamilton_diachronic_2016}, the digital humanities \cite{muralidharan_supporting_2012}, and history \cite{hendrickx_automatic_2011}. 

Our work is inspired by the only previous work on automatically inducing historical sentiment lexicons,
\newcite{cook_automatically_2010};
they use the PMI method and a full modern sentiment lexicon as their seed set,
which relies on the assumption that all these words have not changed in sentiment. 
In contrast, in addition to our different algorithm, we use a small seed set of words that were manually selected based on having strong and stable sentiment over the last 150 years (Table \ref{seeds}; confirmed via historical entries in the Oxford English Dictionary).

\subsection{Examining the lexicons}
\begin{figure*}[t]
\begin{subfigure}{0.5\textwidth}
\centering
\includegraphics[width=0.95\columnwidth]{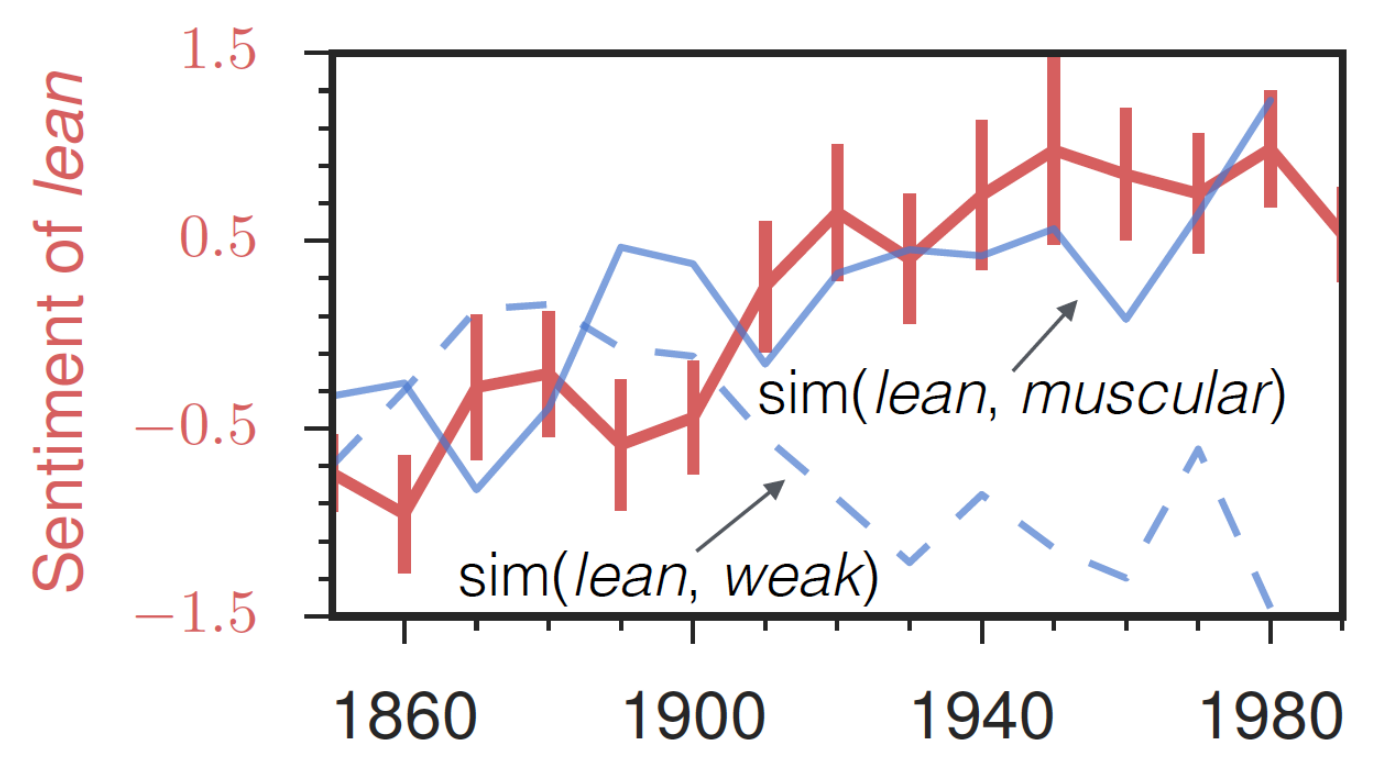}
\caption{{\footnotesize\textbf{\textit{Lean} becomes more positive.} \textit{Lean} underwent amelioration, becoming more similar to \textit{muscular} and less similar to \textit{weak}.}}
\label{lean}
\end{subfigure}
\begin{subfigure}{0.5\textwidth}
\centering
\includegraphics[width=0.95\columnwidth]{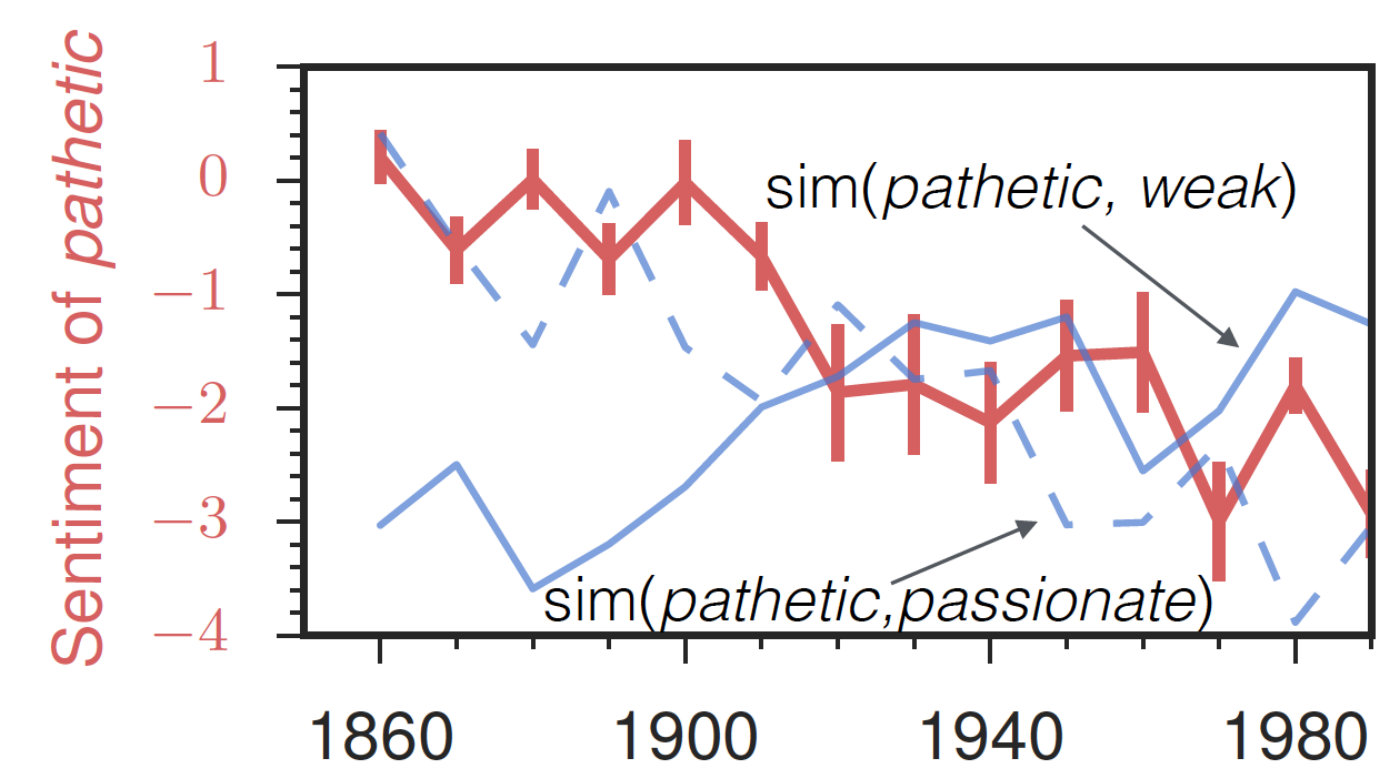}
\caption{{\footnotesize\textbf{\textit{Pathetic} becomes more negative.} \textit{Pathetic} underwent pejoration, becoming similar to \textit{weak} and less similar to \textit{passionate}.}}
\label{pathetic}
\end{subfigure}
\caption{{\footnotesize \textbf{Examples of amelioration and pejoration.}}}
\vspace{-10pt}
\end{figure*}

We constructed lexicons from COHA, since it was carefully constructed to be genre balanced (e.g., compared to the Google N-Grams; \citealt{pechenick_characterizing_2015}). 
We built lexicons for all adjectives with counts above 100 in a given decade and also for the top-5000 non-stop words within each year. 
In both these cases we found that ${>}5\%$ of sentiment-bearing (positive/negative) words completely switched polarity during this 150-year time-period and ${>}25\%$ of all words changed their sentiment label (including switches to/from neutral).\footnote{We defined the thresholds for polar vs.\@ neutral using the class-mass normalization method and compared scores averaged over 1850-1880 to those averaged over 1970-2000.}
The prevalence of full polarity switches highlights the importance of historical sentiment lexicons for work on diachronic linguistics and cultural change. 

Figure \ref{lean} shows an example amelioration detected by this method: the word \textit{lean} lost its negative connotations associated with ``weakness'' and instead became positively associated with concepts like ``muscularity'' and ``fitness''.   Figure \ref{pathetic} shows an example pejoration, where \textit{pathetic}, which used to be more synonymous with \textit{passionate}, gained stronger negative associations with the concepts of ``weakness'' and ``inadequacy'' \cite{simpson_oxford_1989}. 
In both these cases, semantic similarities computed using our learned historical word vectors were used to contextualize the shifts.

Some other well-known examples of sentiment changes captured by our framework include the semantic bleaching of \textit{sorry}, which shifted from negative and serious (``he was in a sorry state'') to uses as a neutral discourse marker (``sorry about that'') and \textit{worldly}, which used to have negative connotations related to materialism and religious impurity (``sinful worldly pursuits'') but now is frequently used to indicate sophistication (``a cultured, worldly woman'') \cite{simpson_oxford_1989}.
Our hope is that the full lexicons released with this work will spur further examinations of such historical shifts in sentiment, while also facilitating CSS applications that require sentiment ratings for historical text.

\section{Conclusion}

\sentprop\ allows researchers to easily induce robust and accurate sentiment lexicons that are relevant to their particular domain of study. 
Such lexicons are crucial to CSS research, as evidenced by our two studies showing that sentiment depends strongly on both social and historical context.

Our methodological comparisons show that simply combining label propagation with high-quality word vector embeddings can achieve state-of-the-art performance competitive with methods that rely on hand-curated dictionaries, and 
the code package released with this work contains a full benchmark toolkit for this area, including implementations of several variants of \sentprop. 
We hope these tools will facilitate future quantitative studies on the domain-dependency of sentiment. 

Of course, the sentiment lexicons induced by \sentprop\ are not perfect, which is reflected in the uncertainty associated with our bootstrap-sampled estimates. 
However, we believe that these user-constructed, domain-specific lexicons, which quantify uncertainty, provide a more principled foundation for CSS research compared to domain-general sentiment lexicons that contain unknown biases. 
In the future our method could also be integrated with supervised domain-adaption (e.g.,\citealt{yang_putting_2015}) to further improve these domain-specific results.

\section*{Acknowledgements} 
The authors thank P. Liang for his helpful comments.
This research has been supported in part by NSF
CNS-1010921,     
IIS-1149837, IIS-1514268
NIH BD2K,
ARO MURI, DARPA XDATA,
DARPA SIMPLEX,
Stanford Data Science Initiative,
SAP Stanford Graduate Fellowship, NSERC PGS-D,
Boeing,          
Lightspeed,			       
and Volkswagen.  
\vspace{-15pt}
\bibliography{social_sentiment}
\bibliographystyle{emnlp2016}

\end{document}